# LAPX: Lightweight Hourglass Network with Global Context

**Haopeng Zhao[1], Marsha Mariya Kappan[1], Mahdi Bamdad[1], Francisco Cruz[1,2]**
[1]School of Computer Science and Engineering, University of New South Wales, Sydney, Australia
[2]Escuela de Ingeniería, Universidad Central de Chile, Santiago, Chile
Emails: haopeng.zhao@unswalumni.com {m.marsha-mariya-kappan, m.bamdad, f.cruz}@unsw.edu.au

## Abstract

Human pose estimation is a crucial task in computer vision. Methods that have SOTA (State-of-the-Art) accuracy, often involve a large number of parameters and incur substantial computational cost. Many lightweight variants have been proposed to reduce the model size and computational cost of them. However, several of these methods still contain components that are not well suited for efficient deployment on edge devices. Moreover, models that primarily emphasize inference speed on edge devices often suffer from limited accuracy due to their overly simplified designs. To address these limitations, we propose LAPX, an Hourglass network with self-attention that captures global contextual information, based on previous work, LAP. In addition to adopting the self-attention module, LAPX advances the stage design and refine the lightweight attention modules. It achieves competitive results on two benchmark datasets, MPII and COCO, with only 2.3M parameters, and demonstrates real-time performance, confirming its edge-device suitability.

## 1 Introduction

Human pose estimation (HPE) aims to locate key body joints, typically from images or video frames, and supports a wide range of real-world applications, like activity monitoring and human-robot interaction. Deep learning methods have dominated this field since the introduction of DeepPose [Toshev and Szegedy, 2014]. Methods at the top of the accuracy leaderboard generally come with large model sizes and heavy computational costs, which limit their deployment on resource-constrained devices and hinder real-time performance [Sun *et al.*, 2019; Xu *et al.*, 2022].

In recent years, many lightweight models have been proposed to address this problem [Kim and Lee, 2020; Yu *et al.*, 2021; Li *et al.*, 2024a]. However, although some popular designs feature a small number of parameters (model size) and low FLOPs (floating-point operations representing computational cost, commonly regarded as a proxy for inference speed), they are not actually fast on edge devices such as smartphones, laptops, or robots without advanced GPUs [Yu *et al.*, 2021; Li *et al.*, 2022a]. Other lightweight models gain popularity through their high inference speed on these devices but achieve only moderate accuracy, as their architectures emphasize simplicity [Sandler *et al.*, 2019; Zhang *et al.*, 2019].

Motivated by the demand for both high accuracy and fast speed in real-time indoor detection, LAP (Lightweight Attention based Pose estimation network) [Kappan *et al.*, 2024] was proposed, a two-stage Hourglass [Newell *et al.*, 2016] network that incorporated efficient convolutions and lightweight attention modules. Please note that in this paper, *hourglass* does not refer to the *Hourglass* network as a whole, but to the fundamental module that composes Hourglass-type networks.

The Hourglass family represents the multi-stage approach, where adding more stages can further improve performance. Convolution-based methods, in general, rely on stacking multiple layers to enlarge the receptive field—the portion of the input that a single neuron can observe—which is crucial for capturing holistic body representations. Since LAP stacks only two stages and is an extremely lightweight version of Hourglass with 2.34M parameters, the most promising directions for improvement are stage design and receptive field enhancement.

Hence, we explored stage design, aiming a better trade-off between the capacity of a single stage and the performance gains from stacking multiple stages. Inspired by Transformer's [Vaswani *et al.*, 2017] extraordinary capacity for modeling long-range dependencies, we apply Non-Local [Wang *et al.*, 2018] modules to capture relationships between distant pixels, enabling the model to achieve a larger receptive field.

By further integrating the soft-gated residual connec-tion [Bulat *et al.*, 2020] and refining other lightweight attention modules, we propose LAPX (**L**ightweight **A**ttention based **P**ose estimation network e**X**tended). It achieves strong results on two benchmark datasets with only 2.3M parameters: 88.6 PCKh@0.5 on MPII [Andriluka *et al.*, 2014] and 70.6 / 72.1 AP on COCO [Lin *et al.*, 2014] at input sizes of $256 \times 192$ and $384 \times 288$, respectively. Although the FLOPs are relatively high (over 2.90G), the model achieves real-time performance of over 15 FPS on an Apple M2 CPU.

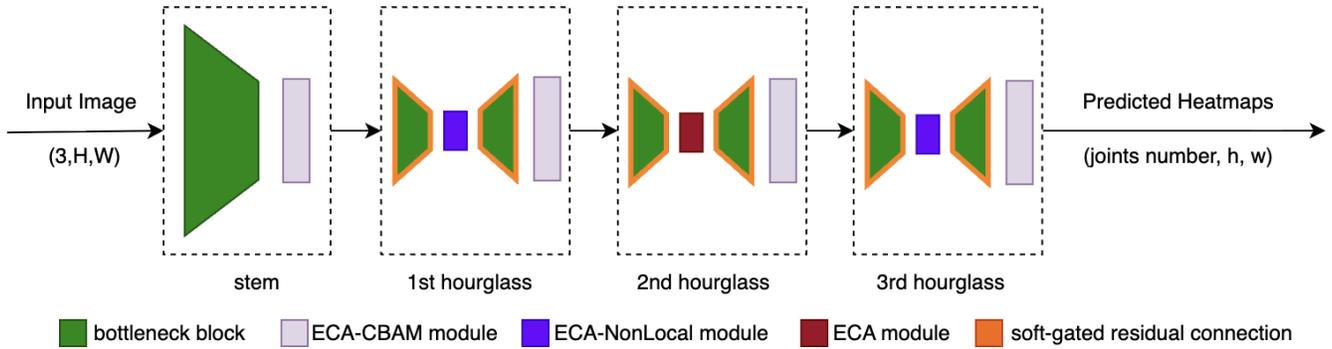

Figure 1: Overall architecture of LAPX. The heights of the blocks represent the spatial sizes of the feature maps. The input RGB image is first processed by the stem, producing initial feature maps while reducing the resolution to the target heatmap size (typically one quarter of the input resolution). This initial feature maps is then sequentially fed through three stacked hourglass modules, finally become the final predicted heatmaps.

The contributions are summarized as follows:

- We identify a better balance between the capacity of a single hourglass module and the benefits of multi-stage stacking under a fixed computational budget.
- We propose the ECA-NonLocal module to introduce global context, together with a tailored training strategy to ensure its effectiveness.
- We design LAPX by replacing the CBAM modules in LAP with ECA-CBAM modules and incorporating the soft-gated residual connection.
- We provide, to the best of our knowledge, the most comprehensive cross-family inference speed comparison of lightweight HPE models—including multi-stage, multi-branch, baseline-like, and transformer-based designs—under a unified edge-device hardware setup (non-GPU), which can serve as a useful reference for future research on practical deployment.

## 2 Related Works

**Heavy methods.** Since DeepPose [Toshev and Szegedy, 2014] first introduced deep learning to HPE, many powerful and heavy deep learning methods have been proposed, revolutionizing the field and largely replacing traditional hand-crafted approaches. In 2D pose estimation, convolution-based methods still dominate, while transformer-based methods have become an increasingly strong force. Convolution-based approaches can be grouped into three categories: multi-stage [Newell *et al.*, 2016; Su *et al.*, 2019; Li *et al.*, 2019], multi-branch [Yang *et al.*, 2017; Sun *et al.*, 2019], and baseline-like [Xiao *et al.*, 2018]. High-Resolution Network (HRNet) [Sun *et al.*, 2019], a representative multi-branch architecture, demonstrated the superiority of maintaining high-resolution branches throughout the entire network, making the multi-branch family the strongest among convolution-based approaches and competitive with recent transformer-based methods in accuracy [Yang *et al.*, 2021; Yuan *et al.*, 2021; Xu *et al.*, 2022].

**Lightweight techniques.** Efficient convolutions have become one of the most popular approaches to lightweight design, as they allow heavy models to drastically reduce parameters while largely preserving architectural strengths. Depthwise separable convolution [Howard *et al.*, 2017], which decomposes standard convolution into depthwise and pointwise components, is the most widely adopted. ShuffleNets [Zhang *et al.*, 2018] introduced group convolution and channel shuffle to further reduce computation. Lightweight attention modules [Hu *et al.*, 2018; Woo *et al.*, 2018; Wang *et al.*, 2020; Li *et al.*, 2022b] have gained recognition for adding minimal computational overhead while improving accuracy, and thus frequently appear in lightweight models to mitigate accuracy loss from parameter reduction.

We compared ShuffleNet-style convolution and depthwise separable convolution as the backbone and found that, under 1.35 GFLOPs, the practical speed of the former was nearly the same as that of the latter at 3.5 GFLOPs. Consequently, we adopted the more hardware-friendly depthwise separable convolution. In addition, we carefully selected lightweight attention modules better suited for LAPX.

**The dilemmas of lightweight methods.** Building on these techniques, light variants of heavy models have been proposed. The evaluation of lightweight models typically focuses on accuracy, model size (parameter count), and computational cost (FLOPs). Although FLOPs are often regarded as a proxy for inference speed, they do not reliably align with actual latency on edge devices.

Kim et al. [Kim and Lee, 2020] presented the first lightweight Hourglass, showing the potential of the multi-stage approach. However, stacking multiple stages greatly increases FLOPs, which may explain why recent lightweight methods rarely adopt this strategy.

Yu et al. proposed Lite-HRNet [Yu *et al.*, 2021], which uses fewer than 0.8 GFLOPs and 2M parameters yet preserves the overall HRNet architecture through a careful design. This extremely low FLOP count established a milestone in the lightweight domain and inspired many subsequent works [Li *et al.*, 2022a; Li *et al.*, 2024b]. Nonetheless, skepticism remains about the practical speed of Lite-HRNet, as reflected in several GitHub discussions. Zhang et al. [Zhang *et al.*, 2024] provided an academic discussion of this issue. While we do not deny the contributions of Lite-HRNet, the extensive use of group convolution and channel shuffle [Zhang *et al.*, 2018] makes it less suitable for current edge devices.

Multi-branch methods are also reported to underperform single-branch designs in real-world speed. Zhang et al. [Zhang *et al.*, 2019] showed that although the single-branch Simple-152 [Xiao *et al.*, 2018] requires more FLOPs than the multi-branch HRNet-W48, Simple-152 achieves faster inference in practice.

Transformer-based methods have also attempted lightweight designs [Yuan *et al.*, 2021; Li *et al.*, 2024a]. However, it is well known that they involve numerous tensor transformations, which are unfavorable for real-time efficiency. To the best of our knowledge, no lightweight transformer-based model has reported non-GPU inference speed. In this study, we provide an empirical measurement of HRFormer [Yuan *et al.*, 2021] for horizontal comparison.

The MobileNets series [Sandler *et al.*, 2018] and Lightweight Pose Network (LPN) [Zhang *et al.*, 2019] can be broadly regarded as baseline-like lightweight models, which have demonstrated good speed on edge devices. However, due to their simple architectures—single branch, single stage, and no advanced modules—their accuracy remains limited under strict parameter budgets.

**LAP [Kappan *et al.*, 2024].** To address the gap between accuracy and real-time performance on edge devices, LAP was proposed. It is an ultra-light Hourglass variant equipped with lightweight attention modules. Although stacking stages results in relatively high FLOPs, its real-time performance on edge devices is good due to its single-branch architecture and hardware-friendly depthwise separable convolutions. Considering the computational budget, LAP was conservatively designed with only two stages, which may limit its ability to fully leverage multi-stage benefits. Like other small convolution-based methods, it also suffers from shallow depth and thus a limited effective receptive field. The design of LAPX was motivated by addressing these two limitations.

**Multi-stage strategy.** The multi-stage strategy was first introduced for iterative refinement in DeepPose. Adding more stages also progressively expands the receptive field. [Newell *et al.*, 2016] showed that later stages improve occluded keypoint prediction, while Chu et al. [Chu *et al.*, 2017] observed that they capture a broader context around joints. Newell et al. presented a detailed ablation study examined stage design by varying the number of stages while keeping the total parameters roughly constant. We adopted the same methodology in our study. However, while their focus was on demonstrating the benefits of additional stages, our emphasis is on identifying the optimal trade-off between the capacity of a single hourglass module and the gains from stacking more stages under an extremely constrained parameter budget.

**Non-Local.** The Non-Local module [Wang *et al.*, 2018] was proposed to overcome the limitation that CNNs (Convolutional Neural Networks) require stacking many layers to capture long-range dependencies. In [Li *et al.*, 2022c], Non-Local modules were fused into an Hourglass model for stereo matching. We also introduce Non-Local modules at the necks of hourglass modules to obtain global context. Different from their work, we introduce ECA-NonLocal modules by placing an ECA [Wang *et al.*, 2020] module before the Non-Local module to enhance channel-wise representation before global dependency modeling in the spatial dimension. Furthermore, given the training instability of Non-Local modules, we propose a sound training strategy.

**Soft-gated residual connection.** The soft-gated residual connection was introduced in [Bulat *et al.*, 2020] to improve information flow in pose estimation networks by dynamically modulating the residual signal. We adopt this design in our work and observe positive results.

## 3 Methodology
### 3.1 Overall architecture

Figure 1 shows the overall model architecture consists of a stem and three hourglass modules. Figure 2 shows the components of a single hourglass module.

The backbone is built with depthwise separable convolutions [Howard *et al.*, 2017] augmented by multiple attention modules.

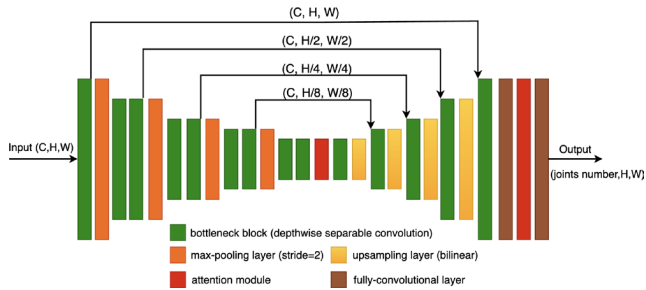

Figure 2: A single hourglass module. This corresponds to any dashed-boxed hourglass in Figure 1, showing all internal components. The width and height of each rectangle represent the channel number and spatial resolution of the feature maps, respectively. All resolutions use the same channel number.

### 3.2 Multi-stage strategy

The advantages of the multi-stage approach arise from two aspects. First, it enables iterative refinement: subsequent stages correct and refine the predictions of preceding ones, leading to progressively higher accuracy. Second, each hourglass module—after multiple downsampling and upsampling steps—produces feature maps that contain both low-level details from the highest resolution and high-level semantics from the lowest resolution. Stacking more stages therefore accumulates more high-level information and effectively enlarges the receptive field.

We fix the total parameter budget of the model and evaluate different stage configurations. For example, a two-stage design uses 256 channels, a three-stage version uses 208 channels, and a four-stage version uses 180 channels. Based on this study, the three-stage design is adopted.

### 3.3 ECA-NonLocal module

Feature maps are three-dimensional ($C \times H \times W$), with the batch dimension omitted. Self-attention can be applied along both the channel ($C$) and spatial ($H \times W$) dimensions, but its quadratic complexity $O(n^2)$—with $n=C$ or $n=H \times W$—is prohibitive for hundreds of channels or high spatial resolutions [Fu *et al.*, 2019]. To keep computation manageable, we apply self-attention only along the spatial dimension at the lowest resolution (e.g., $4 \times 4$), and pair it with a lightweight channel attention module.

As illustrated in Figure 2, the highest-resolution feature maps are downsampled by four successive max-pooling operations until the lowest resolution is reached. Each pixel at this level corresponds to a region in the highest resolution or to a semantic concept such as a hand or chest. In this context, applying self-attention to model the relationships between every pixel and all others can be viewed as establishing all-to-all connections across different regions of the highest-resolution feature maps or exploring the intrinsic relationships among high-level semantics. And thus, the network has a larger effective receptive field.

Our *ECA-NonLocal* module consists of a channel attention module and a spatial attention module in sequence. The channel attention module follows the lightweight ECA design [Wang *et al.*, 2020], but unlike the original, we employ both global average pooling and global max pooling instead of only global average pooling to construct the channel descriptor, followed by a 1D convolution with kernel size *k*=7. The spatial attention module adopts the Non-Local design [Wang *et al.*, 2018] to model all-to-all spatial dependencies. To reduce cost, the query, key, and value embeddings are projected to *C/8* channels instead of *C/2* in the original design, to reduce the parameter overhead and the computational cost. A learnable parameter *γ* is introduce to allow the network to adaptively control how much global context is incorporated.

We found that ECA-NonLocal module is sensitive, showing instability during training. Initially, we incorporated the ECA-NonLocal module into every stage. To identify the best performance, we conducted extensive experiments. We observed that when the model achieved good accuracy, the *γ* values in the first and third stages were positive, while the *γ* value in the second stage was negative. Since *γ* generally reflects the extent to how much global context is utilized, a negative value reduces interpretability. Consequently, we decide to include the ECA-NonLocal module only in the first and third stages.

**Training strategy.** Because global dependencies attend to all locations and are noise-sensitive, which can cause collapse in early epochs before local patterns are established. To ensure stable training and to maximize the performance of the model. Initially, the Non-Local component is frozen so that global information does not interfere too early. After warming up, *γ* gradually increases from 0 to 0.2 over 50 epochs, allowing the network to progressively adapt to global context. Afterwards, *γ* is unfrozen and can be freely learned, allowing the network to decide how much global context it needs.

### 3.4 ECA-CBAM and soft-gated residual connection

We also investigated lightweight attention modules for high-resolution feature maps. We adopt the ECA-CBAM modules instead of the CBAM modules in LAP. Furthermore, we place an additional ECA-CBAM mod-ule immediately after the stem to refine the initial features extracted from the input image, which we refer to as the *stem-ECA-CBAM* module.

Additionally, we introduce the soft-gated residual connection to further improve accuracy by dynamically modulating the residual signal.

## 4 Experiments
### 4.1 Settings

We try our best to ensure reliable results and fair comparisons. For accuracy, we strictly follow the established evaluation protocols [Xiao *et al.*, 2018]. For efficiency, where no unified comparison standard exists, we evaluate models under the same hardware and runtime conditions. We provide our experimental code, configurations and pretrained weights to support further research [1].

**Datasets.** We evaluate our approach on the MPII dataset [Andriluka *et al.*, 2014] and the COCO dataset [Lin *et al.*, 2014], ensuring robustness across benchmarks with different characteristics. MPII contains about 25k images, including ~22k for training, 3k for validation, and 7k for testing. COCO provides ~118k training images and 5k validation images for human pose estimation, with a test set of ~20k images.

**Loss Function.** We adopt mean squared error (MSE) between predicted and ground-truth heatmaps with a visibility mask:

$$L^{(s)} = \frac{1}{2N} \sum_{i=1}^{N} w_i \| H_i^{pred(s)} - H_i^{gt} \|_2^2, \quad (4.1)$$

where $N$ is the number of keypoints, $w_i \in \{0, 1\}$ where 0 masks invisible or out-of-bound joints and $\| \cdot \|_2^2$ denotes the squared $L_2$ norm, that is, the sum of squared elements. Intermediate supervision is applied to three stages, and the final loss is the average:

$$L_{\text{total}} = \frac{1}{3} \sum_{s=1}^{3} L^{(s)}. \quad (4.2)$$

**Metrics.** For MPII, we report the standard metric, *PCKh@0.5*: a keypoint is correct if its error is below 0.5 of the head segment length. For COCO, we follow the official OKS-based protocol, where OKS (Object Keypoint Similarity) measures similarity between predicted and ground-truth poses. We report the standard average precision and recall scores, *AP* (mean AP scores over 10 OKS thresholds: $0.50, 0.55, \ldots, 0.95$).

**Training.** All experiments were conducted on the AutoDL platform[2] with PyTorch 2.5.1 and Python 3.12 on Ubuntu 22.04. The hardware setup consisted of a single NVIDIA RTX 5090 GPU (32 GB), an Intel® Xeon® Platinum 8470Q CPU (25 vCPUs), and 90 GB RAM. CUDA 12.4 was used for GPU acceleration.

We trained with a batch size of 32 using Adam optimizer [Kingma and Ba, 2014]. For MPII, the initial learning rate is $3 \times 10^{-4}$ and is halved at epochs 105, 150, 175, and 190, for a total of 200 epochs. For COCO, the initial learning rate is $2 \times 10^{-4}$ and is multiplied by 0.4 at epochs 120, 160, and 190, for a total of 210 epochs. Augmentation pipeline: Each person instance is cropped from its ground-truth bounding box, randomly scaled, rotated within $[-40°, 40°]$, and horizontally flipped.

**Testing.** For COCO, a pre-trained Faster R-CNN [Ren *et al.*, 2016] human detector provided by SimpleBaseline [Xiao *et al.*, 2018] is used, while for MPII we use the provided ground-truth person boxes.

Three post-processing techniques are applied: (i) a quarter-pixel offset to reduce heatmap quantization error, (ii) flip test with left–right joint correction for robustness, and (iii) heatmap shift (with flip test) to correct spatial misalignment.

Table 1: Comparison on the MPII validation set. All input sizes are $256 \times 256$. TTA denotes test-time augmentation (flip test + heatmap shift).

| Method | Backbone | TTA | #Params | FLOPs | PCKh@0.5 |
|---|---|---|---|---|---|
| **Lightweight HRNet family** | | | | | |
| Lite-HRNet | Lite-HRNet-30 | Yes | 1.8M | 0.42G | 87.0 |
| Dite-HRNet | Dite-HRNet-30 | Yes | 1.8M | 0.40G | 87.6 |
| HF-HRNet | HF-HRNet-18 | Yes | 4.6M | 0.90G | 88.0 |
| HF-HRNet | HF-HRNet-30 | Yes | 7.4M | 1.50G | 88.5 |
| EL-HRNet | EL-HRNet-w32 | No | 5.0M | 2.66G | 87.7 |
| **Lightweight Transformer family** | | | | | |
| LMFormer | LMFormer-L | No | 4.1M | 1.90G | 87.6 |
| **Lightweight Hourglass family** | | | | | |
| LAP | Hourglass | Yes | 2.34M | 3.70G | 87.6 |
| LAPX**(ours)** | Hourglass | No | 2.30M | 3.45G | 88.0 |
| LAPX**(ours)** | Hourglass | Yes | 2.30M | 3.45G | 88.6 |

## 5 Results
### 5.1 Comparisons with Existing Methods

Table 1 summarizes the performance of representative lightweight methods on the MPII validation set. Table 2 presents the results on the COCO validation set, and also reports the actual inference speeds and memory consumptions.

The selected baselines cover both classic and recent approaches. Early work includes LPN [Zhang *et al.*, 2019], proposed in 2019, while more recent models such as HF-HRNet [Zhang *et al.*, 2024], EL-HRNet [Li *et al.*, 2024b], LMFormer [Li *et al.*, 2024a], and LAP [Kappan *et al.*, 2024] were introduced in 2024, ensuring fair temporal relevance in comparison.

Few baselines use ImageNet [Deng *et al.*, 2009] pre-training, and few do not employ test time augmentation

---
[1] https://github.com/Yigemanhahahaha/LAPX
[2] https://www.autodl.com

Table 2: Comparison on the COCO validation set. Pretrain refers to ImageNet pretraining, and TTA stands for test-time augmentation, specifically the combination of flip test and heatmap shift. i7-8700K CPU denotes the Intel Core i7-8700K CPU (3.70 GHz × 12); all inference time results on this device are reported by [Zhang *et al.*, 2019]. All inference time results on the NPU (RK3588) are reported by [Zhang *et al.*, 2024]. Inference times on APPLE M2 CPU are measured by us. RAM usage was measured on our Apple M2 CPU. A dash (–) indicates results that are not reported or are not tested.

| Method | Backbone | Pretrain | TTA | Input | #Params | FLOPs | Inference time | RAM | AP |
|---|---|---|---|---|---|---|---|---|---|
| Lightweight SimpleBaseline family (baseline-like) | | | | | | | | | |
| LPN [Zhang *et al.*, 2019] | ResNet-50 | No | Yes | 256×192 | 2.90M | 1.0G | 17FPS (i7-8700K) ¥ ~35FPS (APPLE M2 CPU) | ~51.5MB | 69.1 |
| LPN | ResNet-101 | No | Yes | 256×192 | 5.3M | 1.4G | 14FPS (i7-8700K) ¥ ~24FPS (APPLE M2 CPU) | ~69.6MB | 70.4 |
| LPN | ResNet-152 | No | Yes | 256×192 | 7.4M | 1.8G | 11FPS (i7-8700K) ¥ ~17FPS (APPLE M2 CPU) | ~90.9MB | 71.0 |
| Lightweight HRNet family (multi-branch) | | | | | | | | | |
| Lite-HRNet [Yu *et al.*, 2021] | Lite-HRNet-30 | No | Yes | 256×192 | 1.8M | 0.31G | ~1.9FPS (RK3588) | - | 67.2 |
| Lite-HRNet | Lite-HRNet-30 | No | Yes | 384×288 | 1.8M | 0.7G | ~0.8FPS (RK3588) | - | 70.4 |
| Dite-HRNet [Li *et al.*, 2022a] | Dite-HRNet-30 | No | Yes | 256×192 | 1.8M | 0.3G | ~1.3FPS (RK3588) | - | 68.3 |
| Dite-HRNet | Dite-HRNet-30 | No | Yes | 384×288 | 1.8M | 0.7G | ~0.5FPS (RK3588) | - | 71.5 |
| HF-HRNet [Zhang *et al.*, 2024] | HF-HRNet-18 | No | Yes | 256×192 | 4.6M | 0.7G | ~23.1FPS (RK3588) ¥ ~15FPS (APPLE M2 CPU) | ~72.6MB | 69.7 |
| HF-HRNet | HF-HRNet-18 | No | Yes | 384×288 | 4.6M | 1.5G | ~10.3FPS (RK3588) | - | 72.4 |
| HF-HRNet | HF-HRNet-30 | No | Yes | 256×192 | 7.4M | 1.1G | ~14.8FPS (RK3588) | - | 70.8 |
| HF-HRNet | HF-HRNet-30 | No | Yes | 384×288 | 7.4M | 2.5G | ~7.3FPS (RK3588) | - | 73.5 |
| EL-HRNet [Li *et al.*, 2024b] | EL-HRNet-W32 | No | No | 256×192 | 5.0M | 2.0G | - | - | 67.1 |
| Lightweight Transformer family (transformer-based) | | | | | | | | | |
| HRFormer [Yuan *et al.*, 2021] | HRFormer-T | Yes | Yes | 256×192 | 2.5M | 1.3G | ~13FPS (APPLE M2 CPU) | ~83.8MB | 70.9 |
| HRFormer | HRFormer-T | Yes | Yes | 384×288 | 2.5M | 1.8G | - | - | 72.4 |
| LMFormer [Li *et al.*, 2024a] | LMFormer-L | No | No | 256×192 | 4.1M | 1.4G | - | - | 68.9 |
| LMFormer | LMFormer-L | No | No | 384×288 | 4.1M | 3.5G | - | - | 70.5 |
| Lightweight Hourglass family (multi-stage) | | | | | | | | | |
| LAP [Kappan *et al.*, 2024] | Hourglass | No | Yes | 256×192 | 2.34M | 2.78G | ~16FPS (APPLE M2 CPU) | ~76.1MB | 72.1 |
| LAP | Hourglass | No | Yes | 384×288 | 2.34M | 6.24G | - | - | 70.9 |
| LAPX**(ours)** | Hourglass | No | No | 256×192 | 2.30M | 2.59G | ~30FPS (APPLE M2 CPU) | ~58.0MB | 69.8 |
| LAPX**(ours)** | Hourglass | No | No | 384×288 | 2.30M | 5.82G | - | - | 71.2 |
| LAPX**(ours)** | Hourglass | No | Yes | 256×192 | 2.30M | 2.59G | ~15FPS (APPLE M2 CPU) | ~58.0MB | 70.6 |
| LAPX**(ours)** | Hourglass | No | Yes | 384×288 | 2.30M | 5.82G | - | - | 72.1 |

(TTA, i.e., flip testing and heatmap shifting). Since both techniques substantially improve accuracy, we explicitly indicate their usage in our tables. For fairness, LAPX results are reported under both TTA and non-TTA settings.

Inference time is rarely reported in prior work, and when provided, it is often restricted to a single model or family. To enable a more comprehensive view of real-time performance on edge devices across different lightweight families, we conduct inference-time evaluations under unified hardware and experimental settings. Specifically, we measure the inference times of LPN (baseline-like), HRFormer [Yuan *et al.*, 2021] (transformer-based), HF-HRNet (multi-branch), LAP, and LAPX (multi-stage) on an Apple M2 CPU using ONNX Runtime without additional acceleration. The reported FPS (frames per second) reflects raw forward inference only; other processing steps are excluded. With TTA, an additional forward pass is required, the speed is estimated to be half of that without TTA.

In addition to inference time, we also report RAM (Random Access Memory) consumption of the above models to provide a more holistic assessment of efficiency on edge devices. When measuring, we excluded the Python interpreter overhead, while including the memory footprint allocated by ONNX Runtime for model parameters and intermediate activations during inference.

**Comparison with HRNet family.** Lite-HRNet [Yu *et al.*, 2021] and Dite-HRNet [Li *et al.*, 2022a] are well

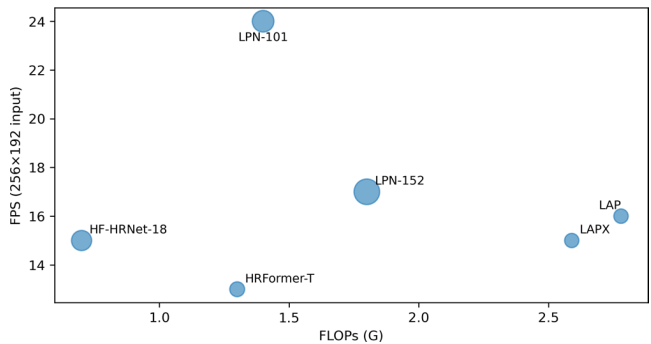

Figure 3: The inconsistency between FLOPs and FPS. The bubble size represents the number of parameters. It can be observed that lower FLOPs do not necessarily correspond to higher FPS, and higher FLOPs do not always lead to lower FPS.

known for their extremely low FLOPs. However, their architectures rely heavily on operations such as group convolution and channel shuffle [Zhang *et al.*, 2018], which are not hardware friendly, as reflected in their limited speed on the RK3588. HF-HRNet was proposed to address this issue.

Our LAPX, with only 2.3M parameters, surpasses the 7.4M HF-HRNet-30 on MPII val and is only 0.2 AP lower on COCO val with 256 × 192 input. At 384 × 288, however, LAPX underperforms the 4.6M HF-HRNet, which we attribute to the limited capacity of 2.3M parameters, preventing full exploitation of richer information at higher resolution.

Although LAPX requires roughly four times the FLOPs of HF-HRNet-18, the actual inference speed is comparable, highlighting that multi-branch design is inherently less efficient than single-branch ones in deployment. Both LAPX and HF-HRNet exhibit relatively low RAM usage (50 MB vs. 70 MB on ONNX Runtime), yet LAPX further reduces the footprint. Therefore, HF-HRNet is more suitable for scenarios that seek higher accuracy using large input image resolutions rather than >15 FPS interaction.

LAPX also consistently outperforms the 5M-parameter EL-HRNet [Li *et al.*, 2024b].

**Comparison with SimpleBaseline family.** Both LPN and LAPX adopt a single-branch architecture, where FLOPs can almost directly reflect the actual inference speed. As expected, LPN demonstrates real-world efficiency consistent with its reputation. LPN ResNet-50 exhibits advantages in both inference time and RAM consumption. However, LAPX surpasses the 5.3M parameter LPN-ResNet-101 by 0.1 AP on COCO val at $256 \times 192$ input resolution. Under the same accuracy pursuit, LPN is faster, while LAPX is smaller.

**Comparison with LAP.** Compared with our previous work LAP [Kappan *et al.*, 2024], LAPX achieves clear improvements on MPII val and COCO val with $384 \times 288$ input, only falls behind on COCO at $256 \times 192$. Overall, LAPX yields higher accuracy. Moreover, LAPX achieves higher accuracy on COCO val when using a larger input resolution, indicating that the additional visual information leads to improved performance, which aligns with common expectations. Considering that LAPX also outperforms LAP on MPII val with a medium input resolution, this further suggests that LAPX exhibits better robustness than LAP.

LAP consumes significantly more RAM than LAPX. This is mainly because the single-stage hourglass modules in LAP employ wider feature channels, resulting in larger intermediate activations. Although LAPX stacks an additional stage, the serial structure allows intermediate activations to be released progressively.

In terms of inference speed, although LAP has marginally higher FLOPs, its actual inference speed is slightly higher than that of LAPX due to the additional lightweight attention modules in LAPX. This observation highlights that, when practical inference speed is critical, models should remain streamlined, with only moderate use of tensor-transforming operations such as attention modules.

**Comparison with Transformer family.** Up to now, lightweight transformer-based methods remain relatively underexplored. LAPX consistently outperforms LMFormer [Li *et al.*, 2024a] in accuracy, which is a purely transformer-based lightweight method.

The strongest accuracy competitor is HRFormer-T, a hybrid lightweight architecture consisting primarily of Transformer layers with a minority of convolution. Leveraging the strengths of Transformer and a multi-branch design, HRFormer-T achieves extremely high accuracy with 2.5M parameters. On COCO val, LAPX is 0.3 AP lower than HRFormer-T at both input resolutions. And, since its framework resembles Swin Transformer [Liu *et al.*, 2021], HRFormer-T also demonstrates practical efficiency, running at 13 FPS on our device.

Although HRFormer-T reports slightly higher accuracy than LAPX, it benefits from additional ImageNet pre-training, whereas LAPX does not. Therefore, the actual performance of LAPX can be considered at least on par with HRFormer-T. This demonstrates that, under an ultra-lightweight regime, a multi-stage convolutional architecture enhanced with global context can absolutely compete with multi-branch and transformer-based designs in terms of accuracy.

On this basis, LAPX also exhibits slightly fewer parameters and marginally faster inference speed, while requiring substantially less RAM than HRFormer-T. Although HRFormer-T contains only 2.5M parameters, its transformer-based design incurs high RAM consumption, nearly matches that of the 7.4M-parameter convolutional model LPN ResNet-152. Consequently, LAPX is positioned as a more practical ultra-lightweight model for industrial deployment.

### 5.2 Ablation study

Table 3 provides the results of the ablation study, highlighting the contribution of each component to the overall performance.

The ablation study is conducted on MPII, which is smaller than COCO and thus allows multiple experiments within an acceptable time and cost.

The total PCKh@0.5 reflects the overall accuracy of the model. The performance on wrist, elbow, knee, and ankle joints can, to some extent, indicate how well the model captures global context. This is because the positions of these joints exhibit greater variability in daily activities and are more prone to occlusion, compared to more stable joints such as the head or hip.

Experiments 5 vs. 6 show that the ECA-CBAM module is more effective than the CBAM module. Comparisons between experiments 2 vs. 7 and 6 vs. 8 highlight the gains introduced by the soft-gated residual connection. Moreover, experiments 6 vs. 9 and 9 vs. 10 demonstrate that inserting an ECA-CBAM module after the stem to refine the initial features consistently improves accuracy by approximately 0.3 PCKh@0.5.

Comparing experiments 1 vs. 2, the three-stage design improves overall PCKh@0.5 by nearly 0.4. The gains are particularly evident for ankle and knee joints, indicating that additional stages enhance the model's ability to capture holistic pose structure. A similar trend appears in experiments 1 vs. 3. However, the four-stage design underperforms the three-stage design. Under a constrained parameter budget, introducing more stages forces each hourglass module to become thinner. Because later stages inherently depend on earlier ones, when the capacity of each individual module is too limited, stacking no longer compensates, which is also reflected in the five-stage design. In general, the three-stage configuration provides the best balance between the capacity of a single stage and the cumulative benefits of stacking.

Table 3: Ablation study on the MPII validation set (256 × 256 input). No test time augmentation is used (no flip test and no heatmap shift). All accuracy values are reported as PCKh@0.5. Note that the total PCKh@0.5 is not the average of the PCKh@0.5 values across individual body parts, since the body parts are not uniformly distributed. SG denotes the soft-gated residual connection. PCKh@0.5 is a very stable evaluation metric. But, as the ECA-NonLocal module led to relatively large performance fluctuations across runs, for the full LAPX model (experiment 12), we conducted three independent runs and report the mean performance, with the standard deviation additionally provided for the total PCKh@0.5.

| Index | Methods | #Params | Head | Shoulder | Elbow | Wrist | Hip | Knee | Ankle | Total |
|---|---|---|---|---|---|---|---|---|---|---|
| | **Multi-stage strategy** | | | | | | | | | |
| 1 | 2-stage (channels=256) | 2.30M | 95.94 | 94.55 | 85.69 | 79.81 | 76.24 | | | 86.45 |
| 2 | *3-stage (channels=208)* | 2.26M | 96.08 | 94.19 | 87.05 | 80.90 | 87.02 | 81.10 | 77.04 | 86.82 |
| 3 | 4-stage (channels=190) | 2.25M | 96.08 | 94.19 | 86.88 | 81.21 | 86.45 | 80.68 | 76.81 | 86.68 |
| 4 | 5-stage (channels=160) | 2.23M | 96.01 | 94.31 | 86.13 | 80.82 | 86.14 | 80.74 | 76.38 | 86.43 |
| | **CBAM vs ECA-CBAM** | | | | | | | | | |
| 5 | 3-stage + CBAM | 2.28M | 96.28 | 94.26 | 87.08 | 81.16 | 86.50 | 81.52 | 77.68 | 86.95 |
| 6 | 3-stage + *ECA-CBAM* | 2.26M | 96.21 | 94.79 | 87.47 | 82.39 | 87.21 | 81.50 | 77.59 | 87.37 |
| | **soft-gated residual connection** | | | | | | | | | |
| 7 | 3-stage + *SG (soft-gated residual connection)* | 2.26M | 96.11 | 94.19 | 87.25 | 81.28 | 87.16 | 81.16 | 77.40 | 87.00 |
| 8 | 3-stage + ECA-CBAM + *SG* | 2.26M | 96.18 | 94.60 | 87.93 | 81.93 | 87.29 | 81.97 | 78.04 | 87.45 |
| | **stem-ECA-CBAM** | | | | | | | | | |
| 9 | 3-stage + ECA-CBAM + *stem-ECA-CBAM* | 2.26M | 96.28 | 94.79 | 88.05 | 82.22 | 87.26 | 82.15 | 78.56 | 87.65 |
| 10 | 3-stage + ECA-CBAM + SG + *stem-ECA-CBAM* | 2.26M | 96.28 | 94.84 | 87.83 | 82.58 | 87.95 | 82.03 | 78.60 | 87.77 |
| | **ECA-NonLocal** | | | | | | | | | |
| 11 | 3-stage + ECA-CBAM + SG + stem-ECA-CBAM + *ECA-NonLocal γ=0.1* | 2.30M | 96.04 | 94.68 | 87.97 | 82.75 | 87.90 | 82.27 | 79.12 | 87.85 |
| 12* | 3-stage + ECA-CBAM + SG + stem-ECA-CBAM + *ECA-NonLocal γ=0.2* | 2.30M | 96.53 | 94.99 | 88.38 | 82.43 | 87.68 | 82.62 | 79.15 | 87.97 ± 0.06 |
| 13 | 3-stage + ECA-CBAM + SG + stem-ECA-CBAM + *ECA-NonLocal γ=0.3* | 2.30M | 96.39 | 94.84 | 87.97 | 82.68 | 87.45 | 82.21 | 78.72 | 87.78 |
| 14 | 3-stage + ECA-CBAM + SG + stem-ECA-CBAM + *ECA-NonLocal γ=0.4* | 2.30M | 96.49 | 94.68 | 87.49 | 81.77 | 86.86 | 81.93 | 77.49 | 87.28 |

Finally, comparing experiments 10 with 11–13 shows that introducing the ECA-NonLocal module, which incorporates global context, leads to improved accuracy, with the setting γ = 0.2 yielding the most favorable results. Observing the poor results of Experiment 14, you can see that introducing too much global context at an early stage may even detrimental. Comparing experiments 10 vs. 12. The gains are very pronounced for elbows, ankles and knees, confirming the effectiveness of expanding the receptive field with global context. In deployment, we further observe that global context enhances prediction stability (reducing jitter) and produces keypoints that align more consistently with the human body.

## 6 Conclusion

LAPX is a three-stage lightweight Hourglass model equipped with the proposed ECA-NonLocal module to leverage the global context. LAPX achieves 88.6 PCKh@0.5 on MPII val and 70.6 / 72.1 AP on COCO val with input sizes of 256 × 192 and 384 × 288, respectively, without ImageNet pretraining. With only 2.3M parameters and real-time performance of 15+ FPS on an Apple M2 CPU, LAPX demonstrates highly competitive performance. Through two generations of work, LAP and LAPX, we reclaim a solid position for multi-stage methods in the lightweight domain. We hope that LAPX can serve as a solid baseline for future lightweight HPE research.

**Future work.** Although we have demonstrated that FLOPs do not directly translate into actual device-level inference speed, we believe that the lightweight Hourglass series could still benefit from pursuing lower FLOPs. In addition, we anticipate that future edge devices with stronger computational capacity will enable better compatibility for strong networks.

## Acknowledgments

My sincere thanks go to my family for providing financial support for my studies. I leave here my grandfather's name, Dongchen, who gave me the most companionship in my childhood. May his name be known to someone.